\documentclass{article}

\usepackage{microtype}
\usepackage{graphicx}
\usepackage{subcaption}
\usepackage{multirow}
\usepackage{amssymb}
\usepackage{amsmath}

\usepackage{makecell}

\usepackage{booktabs} 

\usepackage{hyperref}

\usepackage[accepted]{icml2019}

\icmltitlerunning{Adversarial camera stickers: A physical camera-based attack on deep learning systems}

\begin{document}

\twocolumn[
\icmltitle{Adversarial camera stickers: \\ A physical camera-based attack on deep learning systems}




\begin{icmlauthorlist}
\icmlauthor{Juncheng B. Li}{bosch,cmu}
\icmlauthor{Frank R. Schmidt}{bosch}
\icmlauthor{J. Zico Kolter}{bosch,cmu}

\end{icmlauthorlist}

\icmlaffiliation{bosch}{Bosch Center for Artificial Intelligence}
\icmlaffiliation{cmu}{School of Computer Science, Carnegie Mellon University, Pittsburgh, USA}

\icmlcorrespondingauthor{Juncheng B. Li}{billy.li@us.bosch.com}
\icmlcorrespondingauthor{Frank R. Schmidt}{frank.r.schmidt@de.bosch.com}
\icmlcorrespondingauthor{J. Zico Kolter}{zico.kolter@us.bosch.com}

\icmlkeywords{Machine Learning, ICML}
\vskip 0.3in
]



\printAffiliationsAndNotice{}  

\begin{abstract}
Recent work has documented the susceptibility of deep learning systems to adversarial examples, but most such attacks directly manipulate the digital input to a classifier.  Although a smaller line of work considers physical adversarial attacks, in all cases these involve manipulating the \emph{object} of interest, e.g., putting a physical sticker on an object to misclassify it, or manufacturing an object specifically intended to be misclassified. In this work, we consider an alternative question: is it possible to fool deep classifiers, over \emph{all} perceived objects of a certain type, by physically manipulating the camera itself?  We show that by placing a carefully crafted and mainly-translucent sticker over the lens of a camera, one can create universal perturbations of the observed images that are inconspicuous, yet misclassify target objects as a different (targeted) class. To accomplish this, we propose an iterative procedure for both updating the attack perturbation (to make it adversarial for a given classifier), \emph{and} the threat model itself (to ensure it is physically realizable).  For example, we show that we can achieve physically-realizable attacks that fool ImageNet classifiers in a targeted fashion 49.6\% of the time.  This presents a new class of physically-realizable threat models to consider in the context of adversarially robust machine learning. Our demo video can be viewed at: \url{https://youtu.be/wUVmL33Fx54}
\end{abstract}

\section{Introduction}
\label{submission}

\begin{figure}[t]
    \begin{center}
    \includegraphics[height=1.5in]{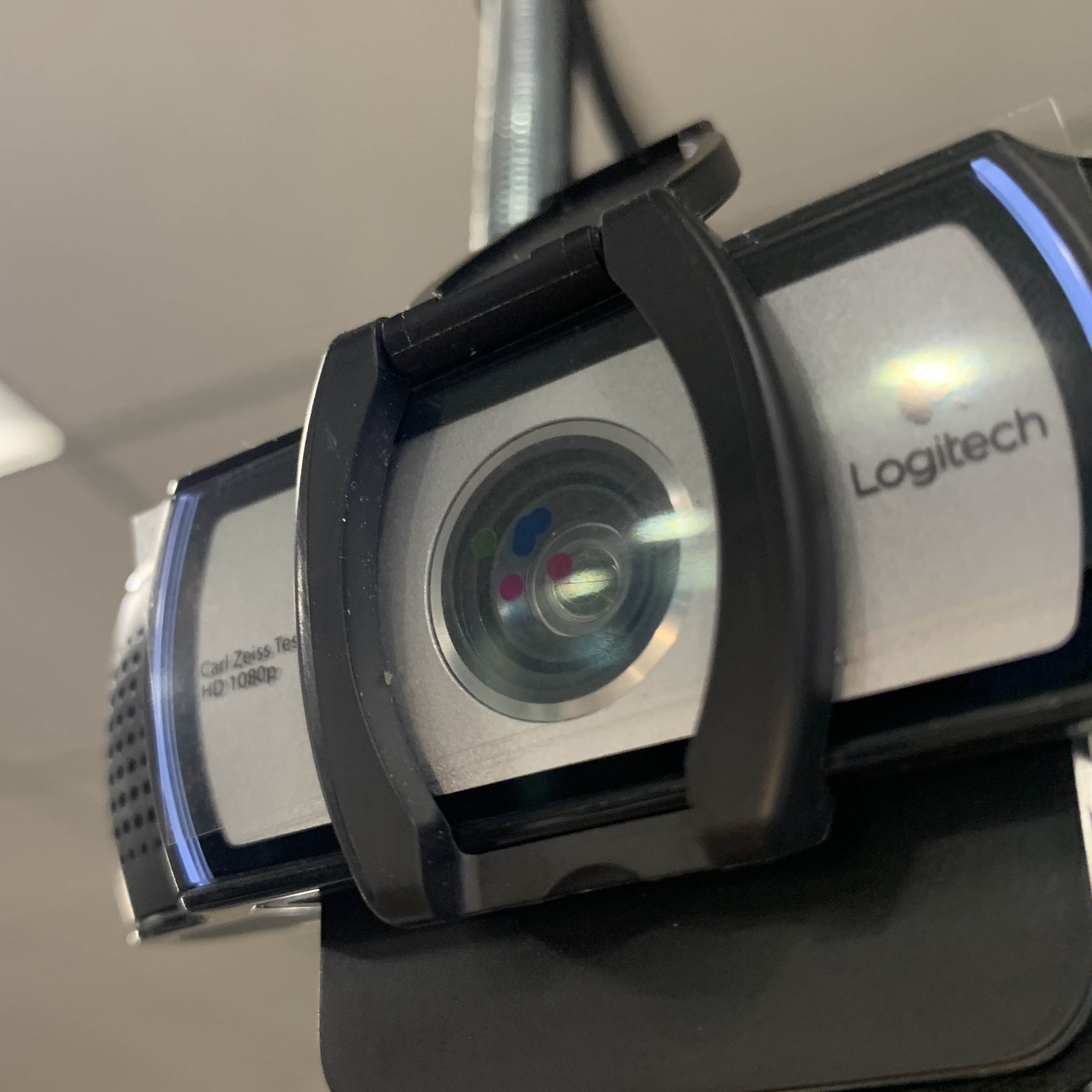}
    \includegraphics[height=1.5in]{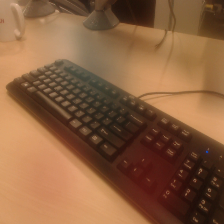}
    \caption{Illustration of our approach: (left) our adversarial sticker affixed to a camera lens; (right) view from the camera with sticker, where the keyboard is reliably classified as a computer mouse for most viewpoints and scales.}
    \label{fig:illustration}
    \end{center}
\end{figure}

Many recent papers have established that deep learning classifiers are particularly susceptible to \emph{adversarial attacks}, manipulations of the input to a classifier specifically crafted to be largely unobservable to humans, but which cause the classifier to predict incorrectly \cite{szegedy2013intriguing,goodfellow2015explaining}. These perturbations be crafted to be bounded by various $L_p$ norms~\cite{su2019one, carlini2017towards, moosavi2016deepfool}. In the majority of the cases studied, however, these attacks are considered in a ``purely digital'' domain, that is they consider perturbations that directly modify the digital input to the classifier in order to fool the model. A smaller but still substantial line of work has emerged to show that these attacks can also transfer to the physical world: these have considered cases of printing out pictures intended to fool classifiers~\cite{brown2017adversarial}, placing physical stickers on objects to fool a classifier \cite{evtimov2017robust}, or directly manufacturing objects intended to fool a classifier (even when observed from multiple angles)~\cite{athalye2017synthesizing}.  In all these cases, though, the primary mode of attack has been manipulating the \emph{object} of interest, often with very visually apparent perturbations. Compared to attacks in the digital space, physical attacks have not been explored to its full extent: we still lack baselines and feasible threat models that work robustly in reality.

In this work, we propose a method for physically fooling a network such that all \emph{objects} of a particular class are misclassified using a visually \emph{inconspicuous} modification.  To accomplish this, we develop an “adversarial camera sticker”, which can be affixed to the lens of the camera itself, and which contains a carefully constructed pattern of dots that causes images viewed by the camera to be misclassified. These dots look like mostly-imperceptible blurry spots in the camera image, and are not obvious to someone viewing the image, as is examplified in Figure~\ref{fig:illustration}. The main challenge of creating this attack relative to past work is that the space of possible perturbations we can feasibly introduce with this model is very limited: unlike past attacks which operate in the pixel-level granularity of the images, the optics of the camera mean that we can only create blurry dots without any of the high frequency patterns typical of most adversarial attacks.  The overall procedure consists of three main contributions and improvements over past work:
\begin{enumerate}
    \item Our threat model is the first of its kind to inject a perturbation on the optical path between the camera and the object, without tampering the object itself.
    
    \item Ours is the first instance we are aware of a ``universal'' physical perturbation (universal perturbations in digital space were considered by \citet{moosavi2017universal}).  Owing to the fact that the sticker will apply the same perturbation to all images, we need to create a single perturbation that will cause the classifier to misclassify over multiple angles and scales.
    
    \item Our method jointly optimizes the adversarial nature of the attack while fitting the threat model to the perturbations achievable by the camera. This is due to the challenge in finding the ``allowable'' set of perturbations that can be physically observed by the camera.
\end{enumerate}

We carefully crafted our dot attacks and the visual threat model (i.e., the set of allowable physical perturbations) using a differentiable alpha blending module (will be explained in Section~\ref{sec:threat_model}). We train the attack, and conducted evaluation experiments, both on a real camera using printed stickers, and (virtually, but with physically realistic perturbations) on the ImageNet dataset~\cite{imagenet_cvpr09}. Our experiments show that on real video data with a physically manufactured sticker, we can achieve an average targeted fooling rate of 52\% over 5 different class / target class combinations, and furthermore reduce the accuracy of the classifier to 27\%.  On the ImageNet test set, we can construct a (digital, but physically realistic) attack that reduces the classification accuracy of the ``street sign'' ImageNet class from 64\% to 18\%, and achieves a average of a 33\% targeted fooling rate over 50 randomly selected target classes.  In total, we believe this work substantially adds to the recent crucial considerations of the ``right'' notion of adversarial threat models (see e.g. \citet{papernot2018sok} for additional discussion on a similar point), demonstrating that these pose physical risk when an attacker can access a camera.

\section{Background and Related Works}

This paper will focus on the so-called white-box attack setting, where we assume access to the model. In this setting, past work can be roughly categorized into two relevant groups for our work: digital attacks and physical attacks.

\subsection{Digital Attacks}
Digital attacks have been relatively well studied since they first rose to prevalence in the context of deep learning in 2014, where \citet{szegedy2013intriguing} used the box-constrained L-BFGS method to find the perturbation. \citet{goodfellow2015explaining} later introduced a simple attack called FGSM which applied perturbation to the direction in image space which yields the highest increase of the linearized cost under $L_\infty$ norm. \citet{kurakin2016adversarial} proposed a iterative multi-step version of this approach, further elaborated upon and studied by~\citet{madry2017towards} as an $L_\infty$ projected gradient descent method.  These approaches, and variants such as those studied independently by \citet{moosavi2016deepfool} (which considers $L_2$ norm attacks), comprise the current state of the art in digital attacks.  

\subsection{Physical Attacks}
Compared with digital attacks, physically realizable attacks have not been explored to a full extent beyond a few existing works. \citet{kurakin2016adversarial} was the first to show the existence of attacks in real world. Soon afterwards, \citet{lu2017no} empirically claimed that such attacks were not easily realized under realistic transformations, though \citet{athalye2017synthesizing} and \citet{evtimov2017robust} soon after demonstrated robust attacks which worked in real world setting. \citet{evtimov2017robust} showed it was possible to modify a stop sign with stickers such that a classifier would classify it as a speed limit sign. However, \citet{lu2017standard} demonstrated that the state-of-the-art detectors are ‘currently’ not fooled by the attacks introduced by \citet{evtimov2017robust}. Often, what works in digital space cannot be generalized to physical space~\cite{zeng2017adversarial}.

Meanwhile, another stream of the works created actual physical objects as attacks.~\citet{brown2017adversarial} created an “adversarial patch,” a small piece of paper that can be placed next to objects to render them all classified as a toaster; \citet{athalye2017synthesizing} manufactured a 3D-printed turtle that is misclassified as rifle or jigsaw puzzle. However, both these attacks may be hard to deploy in some settings because 1) they require that each object of interest for where we want to fool the classifier be explicitly modified; and 2) they are visually apparent to humans when inspecting the object.  

In contrast to the existing methods of physical attacks, our proposed dotted sticker does not require direct tampering of the object of the interest, and it is inconspicuous to the camera viewer. Indeed, it is almost unnoticeable to people looking at the image, unless they specifically know what to look for, as is shown on the left side of Figure~\ref{fig:illustration}.

\section{Crafting Adversarial Stickers}
This section contains the main methodological contribution of our paper: the algorithmic and practical pipeline for manufacturing adversarial camera sticker.  To begin, we will first describe the general threat model (perturbation model) we consider in this work; unlike past works which often considered $L_p$ norm bounded perturbations, we require a threat model that can naturally capture the perturbations seen by a camera.  We describe our approach to fit this perturbation model to observed data (i.e., so that the threat model captures what is actually seen in a given camera), and finally how we adjust the free parameters of our attack to craft adversarial examples.

\subsection{A threat model for physical camera sticker attacks}
\label{sec:threat_model}

Traditional attacks on neural network models work as follows.  Given a classifier $f : \mathcal{X} \rightarrow \mathcal{Y}$, we want to find some perturbation function $\pi: \mathcal{X} \rightarrow \mathcal{X}$, such that for any input $x \in \mathcal{X}$, $\pi(x)$ looks ``indistinguishable'' from $x$, yet is classified incorrectly by $f$ even when $x$ is classified correctly, i.e., $f(x) \neq f(\pi(x))$.  Typically $\pi$ is taken to be some norm-bounded additive perturbation, i.e.,
\begin{equation}
    \pi(x) = x + \delta
\end{equation}
with $\|\delta\|_p \leq \epsilon$ for some bound $\epsilon$ (we generally refer to the space of all such allowable function as $\Pi$), though in this work we will specifically consider a much more limited class of perturbations, to ensure that they are actually achievable in a real system.

The goal of standard adversarial attacks is, for a given $x \in \mathcal{X}, y \in \mathcal{Y}$, to find a perturbation $\pi \in \Pi$ that maximizes the loss
\begin{equation}
    \max_{\pi \in \Pi} \ell(f(\pi(x)),y).
\end{equation}
A slight variant of this approach is to consider \emph{targeted} attacks that specifically try to maximize the loss of the true class and minimize loss of some target class
\begin{equation}
    \max_{\pi \in \Pi} \left (\ell(f(\pi(x)),y) - \ell(f(\pi(x),y_{\mathrm{targ}})) \right ).
\end{equation}
Finally, whereas the standard attacks are able to adapt $\pi$ to a specific input, another variant (which we will need to consider here), is a \emph{universal} perturbation, where a single perturbation function $\pi$ must be chosen to maximize expected error over multiple samples drawn from some distribution $\mathcal{D}$
\begin{equation}
    \max_{\pi \in \Pi} \mathbf{E}_{(x,y) \sim \mathcal{D}} [\ell(f(\pi(x)), y)]
\end{equation}
and where we can also consider the targeted variant as well.

To design a threat model for a physical camera attack, we need to consider the approximate effect of placing small dots on a sticker.  Owing to the optics of the camera lense, a small opaque dot placed upon the camera lens will create a small \emph{translucent} patch on the image itself.  Assuming sufficient lighting, such translucent overlays can be well-approximated by an alpha-blending operation between the original image and an appropriately sized and colored dot.

More formally, explicitly considering $x$ to be an 2D image with $x(i,j)$ denoting the pixel at the $(i,j)$ location, we consider the perturbation function for a \emph{single} dot in the image $\pi_0(x;\theta)$, (where $\theta$ denotes the parameters of the perturbation model, which we will discuss shortly), given by
\begin{equation}
    \begin{split}
    \pi_0(x;\theta)(i,j) & = (1-\alpha(i,j)) \cdot x(i,j) + \alpha(i,j) \cdot \gamma \\
    \alpha(i,j) & = \alpha_{\max} \cdot \exp (-d(i,j)^\beta ) \\
    d(i,j) &= \frac{(i-i^{(c)})^2 + (j-j^{(c)})^2}{r^2}
    \end{split}
\end{equation}
where the parameters:
\begin{equation}
\theta = \left(\gamma, (i^{(c)},j^{(c)}), r, \alpha_{\max}, \beta \right)
\end{equation}
correspond to the following aspects of the dot:
\begin{itemize}
    \item $\gamma \in [0,1]^3$ -- RGB color 
    \item $(i^{(c)},j^{(c)}) \in \mathbb{R}^2$ -- center location (pixel coordinates)
    \item $r \in \mathbb{R}_+$ -- effective radius
    \item $\alpha_{\max} \in [0,1]$ -- maximum alpha blending value
    \item $\beta \in \mathbb{R}_+$ -- exponential dropoff of alpha value
\end{itemize}
Intuitively, this perturbation model captures the following process.  Each dot is parameterized by its center location in the image $(i^{(c)},j^{(c)})$ and its color $\gamma$.  Each pixel $\pi_0(x;\theta)(i,j)$ in the perturbed image is given by a linear combination between the original pixel and the color $\gamma$, weighted by the position-dependent alpha mask $\alpha(i,j)$.  For pixels $(i,j)$ that are closer than the radius $r$ to the point, the corresponding $\alpha(i,j)$ value will be close to $\alpha_{\max}$, whereas for pixels that are far away, the $\alpha(i,j)$ value will be essentially zero.  Finally, the $\beta$ parameters controls the ``smoothness'' of the dropoff: for $\beta \rightarrow \infty$, the alpha mask will hard dropoff at a radius $r$, whereas for smaller values the dropoff in alpha values will be more gradual.

\begin{figure}[t]
\begin{center}
    \includegraphics[width=1.5in]{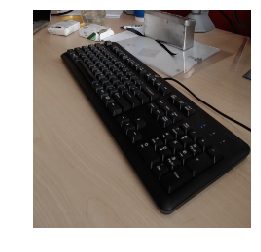}
    \includegraphics[width=1.5in]{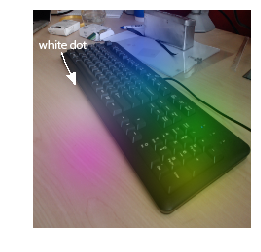}
    \caption{Illustration of perturbation model: (left) original image; (right) image perturbed (digitally) using our alpha-blending perturbation. There are 8 dots in this case, note that the white dot cannot be printed in real world.}
    \label{fig:perturbation-illustration}
\end{center}
\end{figure}

To form our final perturbation model, we simply compose $K$ of these single-dot perturbations, that is
\begin{equation}
    \pi(x;\theta) = \pi_0(\boldsymbol{\cdot} ;\theta_K) \circ \ldots \circ \pi_0(\boldsymbol{\cdot} ; \theta_2)\circ \pi_0(x;\theta_1)
\end{equation}
where the total parameters $\theta = (\theta_1,\ldots,\theta_K)$ are simply the concatenation of the parameters for each dot.  A visualization of a possible perturbation under this model, with multiple dots with different center locations and colors, $\alpha_{\max}$ = 0.3, $\beta = 1.0$, and a radius of 40 pixels is shown in Figure \ref{fig:perturbation-illustration}.  One point that is important to emphasize is that resulting image $\pi(x;\theta)$ is a \emph{differentiable} function of the parameters $\theta$: all parameters are real-valued quantities and each pixel value is a continuous function of the parameters.  In other words, we can implement the perturbation model within an automatic differentiation toolkit (we implement it in the PyTorch library), a feature that will be exploited to both fit the perturbation model to real data, and to construct adversarial attacks, in the following sections.

\subsection{Achieving inconspicuous, physically realizable perturbations}
\label{sec:inconspicuous_pert}

Although the perturbation model above captures a reasonable approximation to attacks that can be created by a physical sticker, there are obvious problems with simply optimizing an attack over its parameters $\theta$.  First, it is easy to create attacks that fool a classifier, but which are very obvious to an observer (consider a fully opaque ``dot'' that covers the entire image).  But second, and more subtly, most parameterizations in such a class will not correspond to perturbation that can be physically achieved on the camera. Indeed, this can be seen in Figure~\ref{fig:perturbation-illustration}: the neon-colored dots are possible to reproduce with a printed sticker, and the precise radius, opacity, and smoothing dropoff of an actual physical dot are not perfectly controllable, but are limited to a very narrow range of achievable parameters. 

Owing to this fact, instead of constructing attacks under the general perturbation model first and then attempting to manufacture them physically, we take an opposite approach: we manufacture physical perturbations that we \emph{know} to be both inconspicuous and (by definition) physically realizable, and we fit the parameters of the perturbation model to recreate these physical perturbations. Ultimately, this will the give us a class of perturbation function $\Pi$ corresponding to \emph{most} of the parameters in our perturbation model being fixed (radius, opacity, dropoff, and a discrete set of allowable colors), while only a few remain free (namely the specific choice of color and the location of the dot).

In more details, our process works like the following. We print a single small physical dot on transparent paper, and collect two images of the same visual scene (with the camera rigidly mounted so as not to move between taking the two images), one with a clean view, referred to as $x^{(0)}$ and one with the dot placed in front of the camera, referred to as $x^{(1)}$; one instance of such images are shown in Figure \ref{fig:diameters}.  We then used the structural similarity (SSIM) \cite{zhao2017loss}, which measures the similarity of 2 pictures, to fit the parameters of our perturbation model to reconstruct the observed perturbation as much as possible; that is, we solve the optimization problem
\begin{equation}
    \max_\theta \; \mathrm{SSIM}(\pi_0(x^{(0)};\theta), x^{(1)})
\end{equation}
where we note that by convention, we want to \emph{maximize} the SSIM to make the images as close as possible.

In theory, because both the SSIM and the perturbation model $\pi$ are differentiable functions, we could simply use projected gradient descent (PGD) to optimize our perturbation model. In reality, however, the loss landscape of this problem is highly non-convex, and most random initializations will have uninformative gradients. Consider the case where the initial guess of dot location does not overlap at all with the actual dot location: in this case, there will be no informative gradient over the dot location (and hence no informative gradients over the other dot parameters either).  In practice, the search procedure therefore requires that we find a very good initialization for the dot location, its color, and the other parameters, and only run PGD from this initial location.  The precise details of our initialization procedure are perhaps somewhat unimportant to the main messages of this paper, but since the practical implementation is a key aspect to this work overall, very briefly, we use the following strategies to find initial points and optimize the SSIM:
\begin{itemize}
    \item To initialize the position we simply make an informed guess based upon the known position of the dot in each such physically perturbed images.
    \item To initialize color space, we fit a linear transformation between the RGB values on the digital version to be printed, and the actual RGB values observed by the camera, that is, we printed 50 separate colors to estimate the transform, $k_{\mathrm{R/G/B}}$ and $b_{\mathrm{R/G/B}}$ are scalars.
    \begin{equation}
      \gamma^{(\mathrm{observed})}_{\mathrm{R/G/B}} = k_{\mathrm{R/G/B}} \cdot \gamma^{(\mathrm{printed})}_{\mathrm{R/G/B}} + b_{\mathrm{R/G/B}}
    \end{equation}
    \item Initial values for the remaining radius, $\alpha_{\max}$, and $\beta$ values are simply fixed to values that produce reasonable-looking images.
    \item In maximizing the SSIM, we found it was advantageous to employ a kind of ``block coordinate descent'' to iteratively optimize over each set of parameters (center location, color, $\alpha_{\max}$, $\beta$, and radius) individually, each using gradient descent.
\end{itemize}

We performed this procedure for 50 different physical dots, to learn a single set of $\alpha_{\max}$, $\beta$, and $r$ parameters, and 50 different colors (and of course, 50 different locations, though these parameters were only fit here in order to fit the remaining parameters well).  After fitting these parameters, we restricted the allowable class of perturbations $\Pi$ to contain up to 10 composed dots, each with fixed $\alpha_{\max}$, $\beta$, and $r$, and with the color gamma coming from a set of 10 possible color choices $\gamma \in \Gamma \equiv \{\gamma^{(1)},\ldots,\gamma^{(10)}\}$, where we chose the 10 colors according to how often a single dot of this color (placed randomly) would cause a ResNet to misclassify an example.  In other words, the ultimate free parameters of our perturbation model are 1) the center location $(i^{(c)},j^{(c)})$ of the dot, and 2) the discrete choice of color $\gamma \in \Gamma$; these are the parameters we subsequently optimize over to create adversarial inputs.

\subsection{Constructing adversarial examples}
\label{sec:real_adverse}

Given this perturbation model, our final goal is to find some $\pi \in \Pi$ (that is, choose center locations and colors) to maximize the fooling rate of the attack.  We specifically consider building a targeted universal adversarial attack against a \emph{single} class $y^\star$, i.e., we want to fool all observed instances of some class $y^\star$ to be labelled as a target class $y_{\mathrm{targ}}$ instead. This is formally specified by the optimization problem
\begin{equation}
    \max_{\pi \in \Pi} \; \mathbf{E}_{x \sim D(x | y^\star)} \left [ \ell(f(\pi(x)),y^\star) - \ell(f(\pi(x)),y_{\mathrm{targ}}) \right ].
\end{equation}
In practice, we approximate this loss by maximizing the loss on some subset of images for a given class.  That is, if $x^{(1)}, \ldots, x^{(M)}$ represent a set of images for class $y^\star$, we maximize the loss
\begin{equation}
    L = \sum_{l=1}^M \left (\ell(f(\pi(x^{(l)})),y^\star) -
   \ell(f(\pi(x^{(l)})),y_{\mathrm{targ}}) \right).
   \label{eq:targeted-universal-loss}
\end{equation}

Again, because the perturbation model and loss are differentiable in the parameters $(i_k^{(c)},j_k^{(c)})$ and $\gamma_k$ (where the subscript $k$ denotes the parameters for the $k$th dot), we could in theory optimize these with pure (projected) gradient descent. However, we here have only a discrete set of allowable color choices $\gamma \in \Gamma$, and as before, the gradients with respect to the dot positions present a highly non-convex loss surface; and unlike in the case of fitting the perturbation model parameters, there is no ``natural'' initialization for these parameters in this highly non-convex space. Owing to this, we resort to a greedy block coordinate descent search over individual block positions and colors.  That is, we discretize the image into a 45 x 45 grid of possible locations for a dot center (45 is chosen such that each is 5 pixels apart for ImageNet), which we denote $\mathcal{C}$, plus the 10 discrete possible colors.  After running coordinate descent in this manner, we finally fine-tune the position of the dots (since the colors are discrete, we cannot fine tune these), using gradient descent over the loss function as well.  The full procedure is shown in Algorithm \ref{alg:secondCoordinateDescent}.

\begin{algorithm}[t]
   \caption{Coordinate Descent: maximize the attack}
   \label{alg:secondCoordinateDescent}
\begin{algorithmic}
   \STATE {\bfseries Input:} Number of dots $K$, True class $y^\star$, target class $y_{\mathrm{targ}}$, dataset $x^{(1)}, \ldots, x^{(M)}$ of images for class $y^\star$
   \STATE {\bfseries Output:} perturbation $\pi \in \Pi$ parametarized by number of dot centers and colors $(i^{(c)}_k, j^{(c)}_k)$ and color $\gamma_k \in \Gamma$ for $k=1,\ldots,K$.
   \STATE {\bfseries Initialization:} $(i^{(c)}_k, j^{(c)}_k) := \emptyset$ (i.e., no visible dot)
   \STATE 
   \STATE // Greedy coordinate descent
   \REPEAT
   \FOR{$k \in 1,\ldots,K$}
   \STATE // Test all locations and colors for this dot \\
   \FOR{$(i^{(c)}_k, j^{(c)}_k), \gamma_k \in \mathcal{C} \times \Gamma$}
   \STATE Evaluate loss \eqref{eq:targeted-universal-loss}
   \ENDFOR
   \STATE Choose $(i^{(c)}_k, j^{(c)}_k), \gamma_k$ to be the parameters that achieved the highest loss.
   \ENDFOR
   \UNTIL Loss converges
   \STATE 
   \STATE // Gradient descent fine-tuning
   \REPEAT
   \FOR{$k \in 1,\ldots,K$}
       \STATE Update $i^{(c)}_k, j^{(c)}_k$ via gradient descent on loss \eqref{eq:targeted-universal-loss}
   \ENDFOR
   \UNTIL Loss converges      
\end{algorithmic}
\end{algorithm}

\section{Experiments}
Here we present experiments of our attack evaluating both the ability of the digital version of the attack to misclassify images from the ImageNet dataset (still restricting perturbations to be in our physically realizable subset), and evaluating the system on two real-world tasks: classifying a computer keyboard as a computer mouse, and classifying a stop sign as a guitar pick.  We also detail some key results in the process of fitting the threat model to real data.

\subsection{Experimental setup}
\begin{figure}[t]
\begin{center}
    \includegraphics[width=1in]{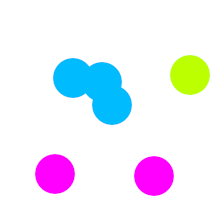}
    \includegraphics[width=1in]{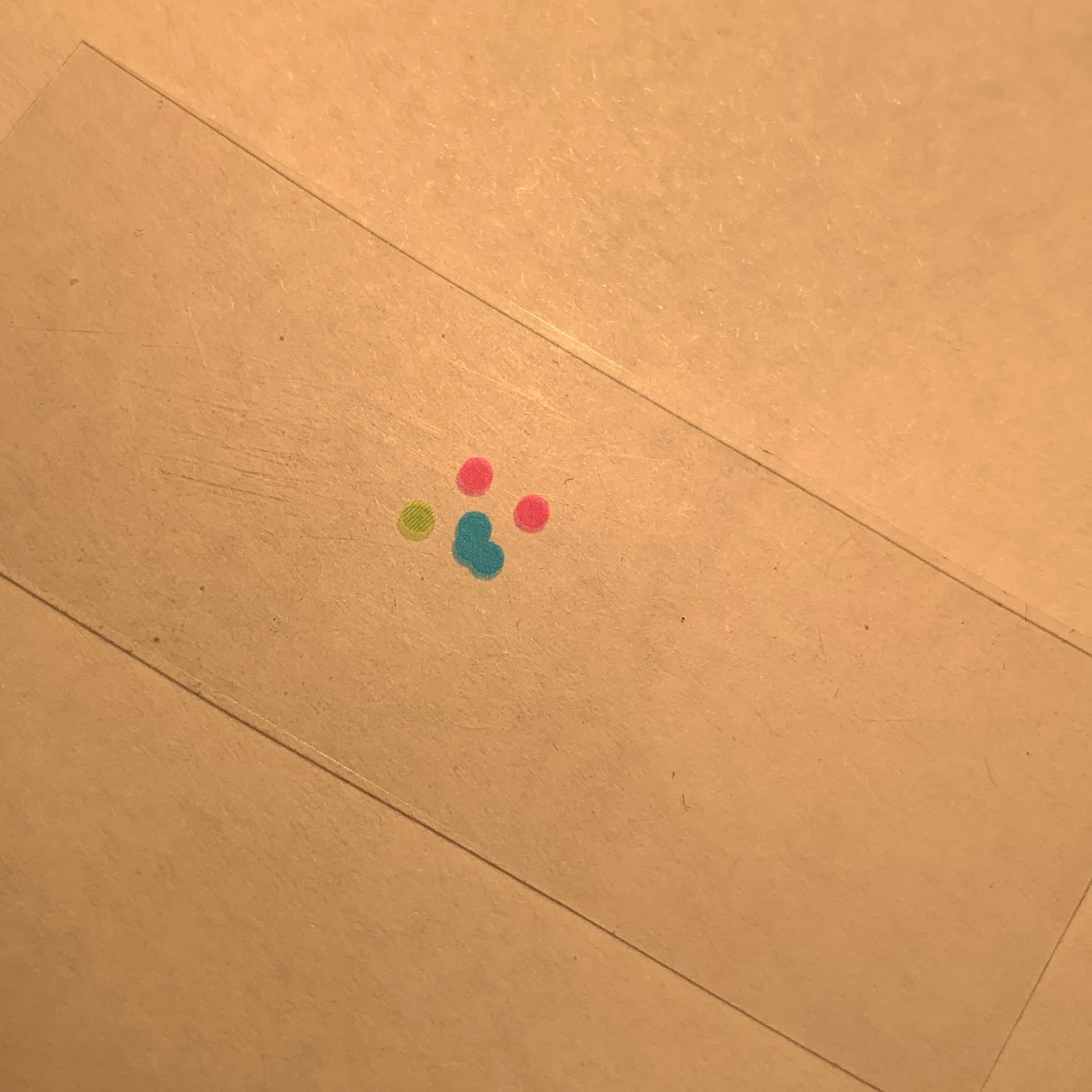}
    \includegraphics[width=1in]{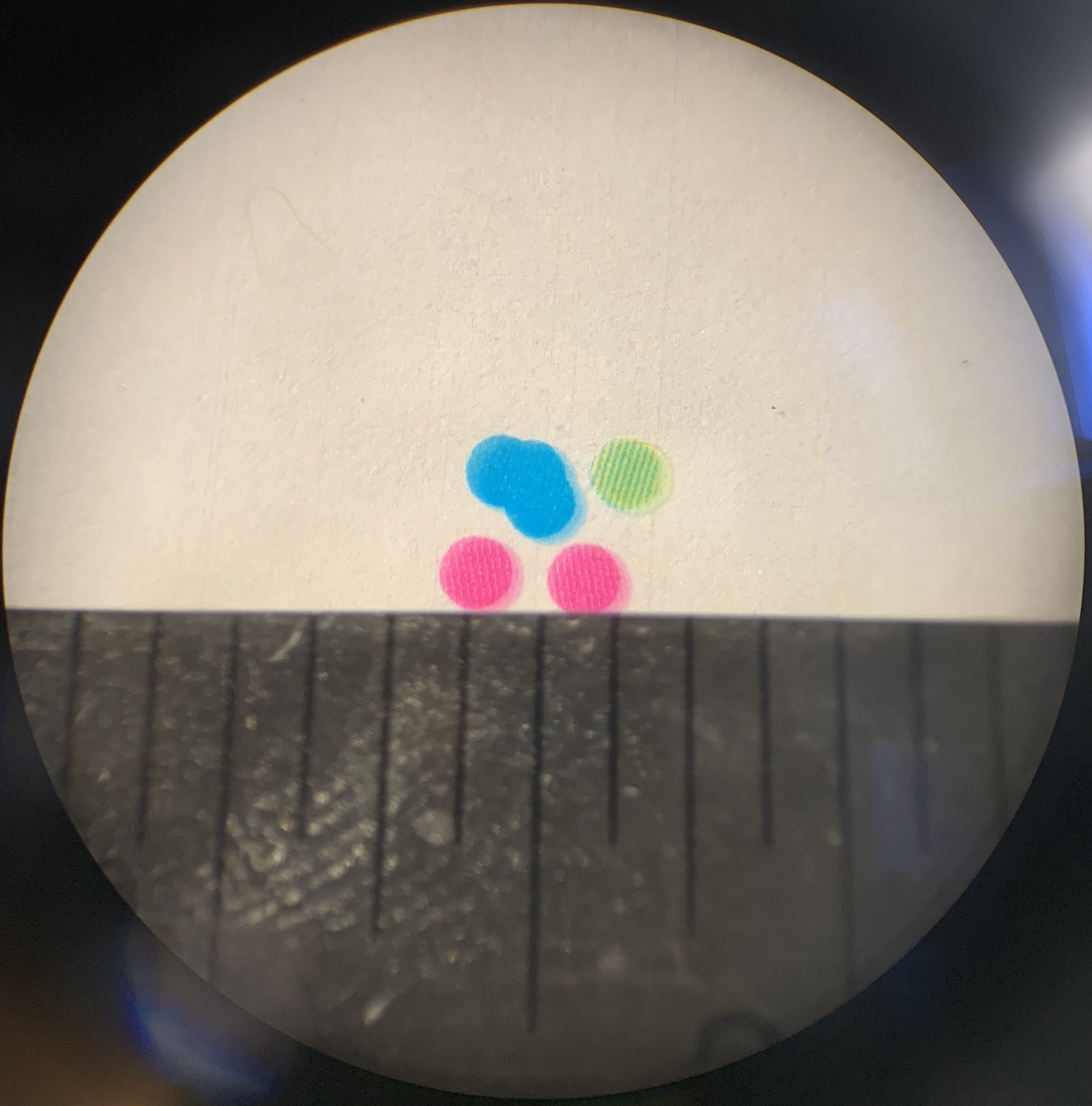}
  \caption{Illustration of $r = 0.025$ inch dots printed on the camera sticker from Photoshop, corresponding to $r = 40 pixel$ dots in camera space: (left) bitmap representation of sticker; (middle) printed sticker on transparency; (right) microscope view.}
  \label{fig:printing}
  \end{center}
\end{figure}

All our experiments consider fooling a ResNet-50 \cite{he2016deep} classifier, pretrained using the ImageNet dataset \cite{imagenet_cvpr09}; we specifically use the pretrained model included in the PyTorch library \cite{paszke2017automatic}.
We use an HP Color Laser Jet M253 to print the dot patterns on transparency papers; both the printer and the paper are off-the-shelf office supplies. We can print different sized dots with radius no smaller than 0.01 inches in multiple colors.  
Figure~\ref{fig:printing} shows the printing pipeline. 

\subsection{Training and Classification on ImageNet}

\begin{figure}[t]
\begin{center}
    \includegraphics[width=1.2in]{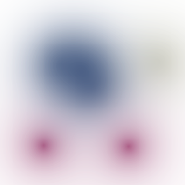}
    \includegraphics[width=1.2in]{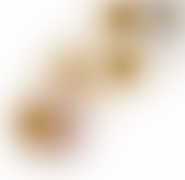}
    \caption{Perturbations (illustrated over a white background) corresponding to the targeted attacks found for the (left) ``computer keyboard'' to ``computer mouse'' and (right) ``street sign'' to ``guitar pick'' adversarial attacks.  In both cases the 6-dot attacks can be physically manufactured, these figures show a digital representation.}
    \label{fig:resulting-perturbations}
\end{center}
\end{figure}

To train and evaluate the system on a broad range of images, we use the same ImageNet dataset used to train the classifier originally.  For example, to train a ``computer keyboard'' to ``computer mouse'' attack, we optimize the parameters of our attack perturbation, using Algorithm \ref{alg:secondCoordinateDescent}, with the 1000 images in the ImageNet dataset corresponding to computer keyboards, and using the computer mouse class as the target class.  Figure \ref{fig:resulting-perturbations} shows the learned perturbations for two instances.  Note that these are still \emph{digital} representations of the perturbations, but they are constrained such that we already know how to manufacture physical realization of these dots, as we will show in the subsequent section; however, all results in this section will be dealing with the digital versions.


\vspace{-0.5cm}
\begin{table}[H]
    \centering
    \caption{Performance of our 6-dot attacks on ImageNet test set.} 
    \label{table:results-imagenet}
    \begin{tabular}{cc|ccc} \toprule
        & & \multicolumn{3}{c}{\thead{Prediction}} \\
         \thead{Class} & \thead{Attack} & \thead{Correct} & \thead{Target} &\thead{Other}  \\ \midrule
         Keyboard $\rightarrow$ & No & 85\% & 0\% & 15\% \\
         Mouse & Yes & 48\% & 36\% & 16\% \\ \midrule
         Street sign $\rightarrow$ & No & 64\% & 0\% & 36\% \\
         Guitar Pick & Yes & 32\% & 34\% & 34\% \\ \midrule
         Street sign $\rightarrow$ & No & 64\% & 0\% & 36\% \\
          50 classes & Yes & 18\% & 33\% & 49\% \\ \midrule
         50 Classes $\rightarrow$ & No & 74\% & 0\% & 26\% \\
         50 classes & Yes & 42\% & 31\% & 27\% \\ \bottomrule
    \end{tabular}
\end{table}

Table \ref{table:results-imagenet} shows the ability of our learned perturbations (6 dots) to fool images from the ImageNet test set for these two categories. We also showed the average success rate to fool stop sign into 50 random classes. More generally, we also include the averaged results for fooling 50 random classes into other 50 random target classes. While our attacks cause smaller increase in error compared to what is common from traditional purely-digital attacks, we emphasize that the allowable class of perturbation is very much smaller in our setting.  In particular, by constraining perturbations to be both visually imperceptible \emph{and} realizable by placing a sticker on the lens, we are significantly constraining the allowable parameter space, and further constraining it to have relatively low-frequency components as compared to traditional attacks.  Thus, the fact that we can decrease the accuracy so substantially with a relatively small overlay is quite notable. Of course, the real test of the method is the ability to transfer these attacks to the real world, as we discuss in the next section.

\begin{figure}[t]
\begin{center}
  \begin{subfigure}[b]{0.15\textwidth}
    \includegraphics[width=\textwidth]{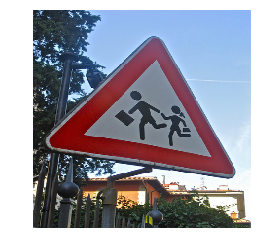}
    \caption{``Street sign"}
    \label{fig:digi_bg_2}
  \end{subfigure}%
  \begin{subfigure}[b]{0.15\textwidth}
    \includegraphics[width=\textwidth]{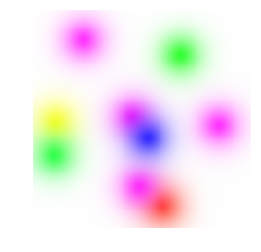}
    \caption{Perturbation}
    \label{fig:digi_opaq_pert}
  \end{subfigure}%
  \begin{subfigure}[b]{0.15\textwidth}
    \includegraphics[width=\textwidth]{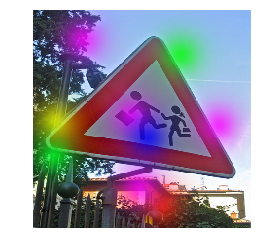}
    \caption{``Projector"}
    \label{fig:digi_opaq_result}
  \end{subfigure}
  \end{center}
  \caption{Physically unrealizable results generated by directly optimizing digital attacks. Here, ``projector" is the model's prediction of the perturbed image of a ``street sign".}
  \label{fig:digi_attack}
\end{figure}

\paragraph{Unrealizable attacks}
\label{sec:digi_attack}
Before moving to the physical attacks, we want to emphasize that the threat model we discuss, if \emph{not} constrained to be physically realizable \emph{or} particularly inconspicuous, can produce much higher fooling rates;
By simple running PGD on the underlying parameters of the threat model within allowable range $\gamma \in [0,1]^3$; $(i^{(c)},j^{(c)}) \in [0,224]^2$; $r \in [1,60]$; $\alpha_{\max} \in [0,1]$; $\beta \in \mathbb{R}_+$, we can indeed get a strong (universal) perturbation, but one which would still be considered ``adversarial'' by the context of much past work (in that a human could still identify the true underlying image).
Figure~\ref{fig:digi_attack} illustrates an additional instances of universal perturbations in digital space that fools ``street sign" into ``projector" which achieves 83\% fooling rate on the ImageNet test set. However, Figure~\ref{fig:digi_opaq_pert} shows an attack being overly opaque. It also requires bright colors that are not possible to print and view via a transparent sticker. Thus, the result mainly serves to emphasize that explicitly fitting the threat model to physical data is crucial for achieving realistic perturbations.

\subsection{Evaluation of attacks in the real world}
\begin{figure}[t]
\begin{center}
    \includegraphics[width=0.4\textwidth]{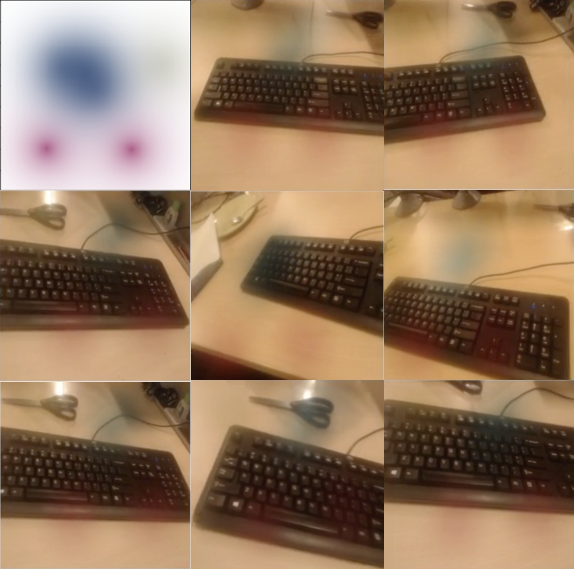}
    \caption{Sticker perturbation to fool ``computer keyboard" class to ``mouse" class}
    \label{fig:9angles}
    \end{center}
\end{figure}
\begin{figure}[t]
\begin{center}
    \includegraphics[width=0.39\textwidth]{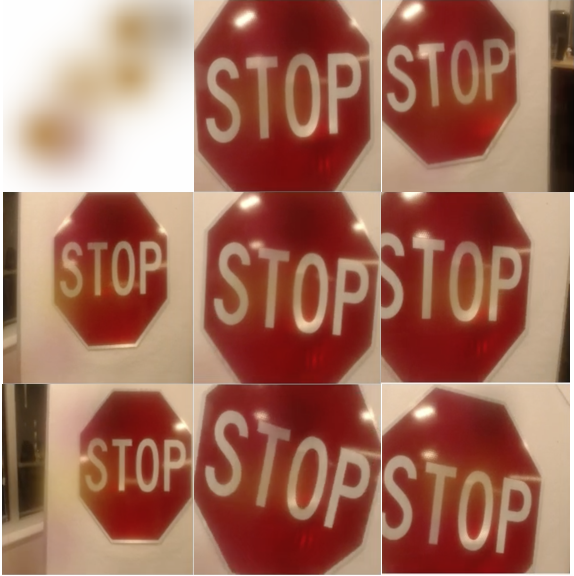}
    \caption{Sticker perturbation to fool ``street sign" class to ``guitar pick" class}
    \label{fig:stopsign}
\end{center}
\end{figure}

In this section we present our main empirical result of the work, illustrating that the perturbations we produce can be adversarial in the real world when printed and applied physically to a camera, and when viewing a target object at multiple angles and scales. 
A video demonstrating the attack is available at  \url{https://youtu.be/wUVmL33Fx54}.
This is the first time that such an adversarial attack has been demonstrated in the real world (that is, an attack that modifies the camera viewing the scene instead of the object), and it opens up a new attack vector for adversarial examples against deep learning systems.

Specifically, using the process above, we printed the physical stickers corresponding to the two attacks mentioned in the previous section, and affixed them to our camera.  We then recorded short videos of the camera viewing these two physical objects at a number of different angles and scales.  In particular, for each case we created 1000 frame video of the camera (with the sticker attached) viewing the target object from multiple angles, and used our ResNet-50 classifier to predict the class of each object.   

Figure \ref{fig:9angles} and \ref{fig:stopsign} shows several snapshots of the process for both the keyboard and stop sign tests.  The overlay created by the adversarial sticker is visible in all cases, but relatively inconspicuous, and would be unlikely to be noticed (or more likely discarded as merely some dust on the camera) by someone not primed to look for the particular patterns.  Table \ref{tab:video_fooling} shows the performance of the ResNet-50 classifier more quantitatively for all 1000 frames of each video.  In this table, we break down the number of frames predicted correctly (as a keyboard or street sign), the number of frames predicted as the target class of the attack (mouse or guitar pick respectively), and the number of frames predicted as some other class.  The inclusion of the target class here (and the fact that the majority of images are predicted as this class) is important: many innocuous transformations to ImageNet-like images, such as rotations, translations, or non-traditional cropping, can fool a classifier \cite{engstrom2017rotation,azulay2018deep}.  However, since we are, the majority of the time, producing the \emph{target} class, we emphasize that this is not merely a manner of random noise fooling the classifier; rather, this specific pattern of dots is tuned precisely to produce the intended target class, and this manifests both in the digital and physical domains.

\begin{table}[t]
    \centering
    \caption{Fooling performance of the our method on two 1000 frame videos of a computer keyboard and a stop sign, viewed through a camera with an adversarial sticker placed on it targeted for these attacks.  The ``correct predictions'' column indicates how many times the object is correct classified; the ``targeted attack prediction'' indicates how often the target class is predicted (computer mouse and guitar pick for the two classes respectively); and the third column indicates how often another class is predicted.} 
    \label{tab:video_fooling}
    \begin{tabular}{c|ccc} \toprule
    & \multicolumn{3}{c}{\thead{Prediction}} \\
    \thead{Class} & \thead{Correct} & \thead{Targeted} & \thead{Other} \\ \midrule
    Keyboard & 271 & \makecell{548 \\ (mouse)}  & 181 \\ 
    Keyboard & 320 & \makecell{522 \\ (space bar)}  & 158 \\ \midrule
    Street sign & 194 & \makecell{605 \\(guitar pick)} & 201 \\
    Street sign & 222 & \makecell{525\\(envelope)} & 253 \\ \midrule
    Coffee mug & 330 & \makecell{427 \\(candle)} & 243 \\\bottomrule
    \end{tabular}
\end{table}

\subsection{Other experiments}

Finally, because there are several aspects of interest regarding both the power of the thread model we consider and our ability to physically manufacture such dots, we here present additional evaluations that highlight aspects of the setup.

\paragraph{Effect of the number of dots}
Although most of the aspects of our dots (besides color and position) are fixed by fitting them to observed data, one free parameter that we \emph{do} have at our disposal is the number of dots we place on the sticker.  Table \ref{tab:final_attack} shows the targeted fooling rate for the computer keyboard to computer mouse attack (here evaluated in the digital realm on the ImageNet test set), varying the number of dots.  As expected, additional dots increases the fooling rate, without reaching diminishing returns even at 10 dots.  Thus, the limiting factor in the attack is largely related to how many dots we can physically print while remaining indistinct; for the physical attacks in the previous section, for this reason we limited the stickers to 6 dots (which also highlights that the fooling rate on these real instances is substantially higher than for the ImageNet test set).

\begin{table}[t]
\begin{center}
\caption{Number of Dots and Targeted Fooling Rate against ResNet-50 model on the ImageNet test set}
\label{tab:final_attack}
\begin{tabular}{cc}
\toprule
\thead{Number of Dots} & \thead{Targeted Fooling Rate} \\
\midrule
1 & 27.9\% \\ 
3 & 32.0\% \\ 
5 & 34.2\% \\ 
7 & 38.2\% \\ 
10 & 49.6\% \\ \bottomrule
\end{tabular}
\end{center}
\end{table}
\vspace{-0.2cm}

\begin{figure}[t]
  \begin{subfigure}[b]{0.16\textwidth}
    \includegraphics[width=\textwidth]{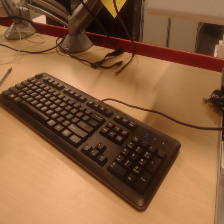}
    \caption{Background}
    \label{fig:trans_bg}
  \end{subfigure}%
  \begin{subfigure}[b]{0.16\textwidth}
    \includegraphics[width=\textwidth]{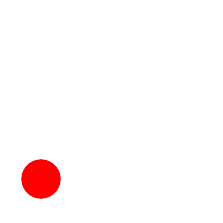}
    \caption{Sticker Pattern}
    \label{fig:1-red}
  \end{subfigure}%
  \begin{subfigure}[b]{0.16\textwidth}
    \includegraphics[width=\textwidth]{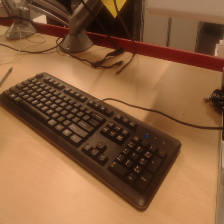}
    \caption{ r = 0.011 inch }
    \label{fig:0022}
  \end{subfigure}
  \begin{subfigure}[b]{0.16\textwidth}
    \includegraphics[width=\textwidth]{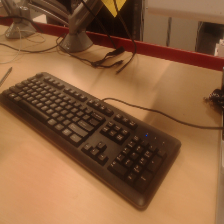}
    \caption{ r = 0.015 inch }
    \label{fig:003}
  \end{subfigure}%
  \begin{subfigure}[b]{0.16\textwidth}
    \includegraphics[width=\textwidth]{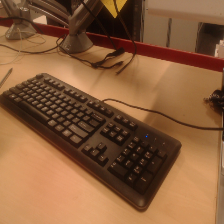}
    \caption{ r = 0.02 inch }
    \label{fig:004}
  \end{subfigure}%
  \begin{subfigure}[b]{0.16\textwidth}
    \includegraphics[width=\textwidth]{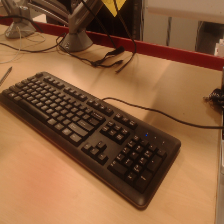}
    \caption{ r = 0.025 inch }
    \label{fig:005}
  \end{subfigure}
  \caption{Resulting effects of a single red dotted stickers with different radius (the dot is applied to the bottom left corner of the camera view) }
  \label{fig:diameters}
\end{figure}

\paragraph{Effect of printed dots on camera perturbations}
Last, although these are largely points of discussion, we highlight some important properties about the effects of the physical printed dots on the perturbations observed by the camera.  It should be noted above that all our physically manufactured stickers, such as those shown in Figure \ref{fig:printing}, are (small) solid opaque dots, which we then place over the camera lens.  A natural question arises as to whether we could create even smaller dots in the perturbation space by simply printing a smaller dot.  However, this does not work: due to the optics of the camera, after a certain point a smaller size printed dot does not result in a smaller size visual dot in the camera, but rather merely a \emph{more transparent} dot of the same size.  This process is shown in Figure \ref{fig:diameters}, and also highlights why we opt to use small solid printed dots to fit our threat model, rather than allowing for the printed dots to be transparent.

\section{Conclusion}
In this paper, we demonstrated that it is possible to adversarially fool deep image classifiers in the real world, not by modifying an object of interest itself, but by modifying the camera observing the objects.  The optics involved with such modifications greatly limits the types of attacks that can be physically realized, but by developing a reasonable threat model and then fitting the parameters of this model to data, we can accurately capture the allowable set of perturbations.  By then optimizing over this class to construct (universal) adversarial examples, we show that it is possible to perform targeted adversarial attacks on real objects, using a single adversarial sticker to misclassify an object in multiple different orientations and scales.  Overall, this suggests a new vector of attack against machine learning algorithms deployed in the real world, highlighting the importance of adversarial robustness from a practical, security-based point of view. 


\clearpage 
\bibliography{main}
\bibliographystyle{icml2019}


\end{document}